\documentclass[wcp, 10pt]{jmlr}

\usepackage{multirow}
\usepackage{multicol}
\usepackage{booktabs}
\usepackage{natbib}

\usepackage{verbatim}
\usepackage[final]{changes}

\newcommand{\bx}{\mathbf{x}}
\newcommand{\NP}{\emph{NeuralPower}\xspace}

\let\cline\cmidrule

\title[]{\NP: Predict and Deploy Energy-Efficient Convolutional Neural Networks}

 \author{\Name{Ermao Cai} \Email{ermao@cmu.edu}\\
 \addr Department of ECE, Carnegie Mellon University, Pittsburgh, PA, USA
 \AND
 \Name{Da-Cheng Juan} \Email{dacheng@google.com}\\
 \addr Google Research, Mountain View, CA, USA
  \AND
\Name{Dimitrios Stamoulis} \Email{dstamoul@andrew.cmu.edu }\\
 \addr Department of ECE, Carnegie Mellon University, Pittsburgh, PA, USA
 \AND
 \Name{Diana Marculescu} \Email{dianam@cmu.edu}\\
 \addr Department of ECE, Carnegie Mellon University, Pittsburgh, PA, USA
}
\editors{Yung-Kyun Noh and Min-Ling Zhang}

\begin{document}

\maketitle

\begin{abstract}
``How much energy is consumed for an inference made by a convolutional neural network (CNN)?'' With the increased popularity of CNNs deployed on the wide-spectrum of platforms (from mobile devices to workstations), the answer to this question has drawn significant attention. From lengthening battery life of mobile devices to reducing the energy bill of a datacenter, it is important to understand the energy efficiency of CNNs during serving for making an inference, before actually training the model. In this work, we propose \NP: a layer-wise predictive framework based on sparse polynomial regression, for predicting the serving energy consumption of a CNN deployed on any GPU platform. Given the architecture of a CNN, \NP provides an accurate prediction and breakdown for power and runtime across all layers in the whole network, helping machine learners quickly identify the power, runtime, or energy bottlenecks. We also propose the \textit{``energy-precision ratio'' (EPR)} metric to guide machine learners in selecting an energy-efficient CNN architecture that better trades off the energy consumption and prediction accuracy. The experimental results show that the prediction accuracy of the proposed \NP outperforms the best published model to date, yielding an improvement in accuracy of up to $68.5\%$. We also assess the accuracy of predictions at the network level, by predicting the runtime, power, and energy of state-of-the-art CNN architectures, achieving an average accuracy of $88.24\%$ in runtime, $88.34\%$ in power, and $97.21\%$ in energy. We comprehensively corroborate the effectiveness of \NP as a powerful framework for machine learners by testing it on different GPU platforms and Deep Learning software tools.

\end{abstract}


\section{Introduction}
\label{sec:introduction}
``How much energy is consumed for an inference made by a convolutional neural network (CNN)?'' ``Is it possible to predict this energy consumption before a model is even trained?'' ``If yes, how should machine learners select an energy-efficient CNN for deployment?'' These three questions serve as the main motivation of this work.

In recent years, CNNs have been widely applied in several important areas, such as text processing and computer vision, in both academia and industry. 
However, the high energy consumption of CNNs has limited the types of platforms that CNNs can be deployed on, which can be attributed to both (a) high power consumption and (b) long runtime. GPUs have been adopted for performing CNN-related services in various computation environments ranging from data centers, desktops, to mobile devices. In this context, resource constraints in GPU platforms need to be considered carefully before running CNN-related applications.

\begin{figure}[ht!] 
	\centering
	\small
	\includegraphics[width=0.9\linewidth]{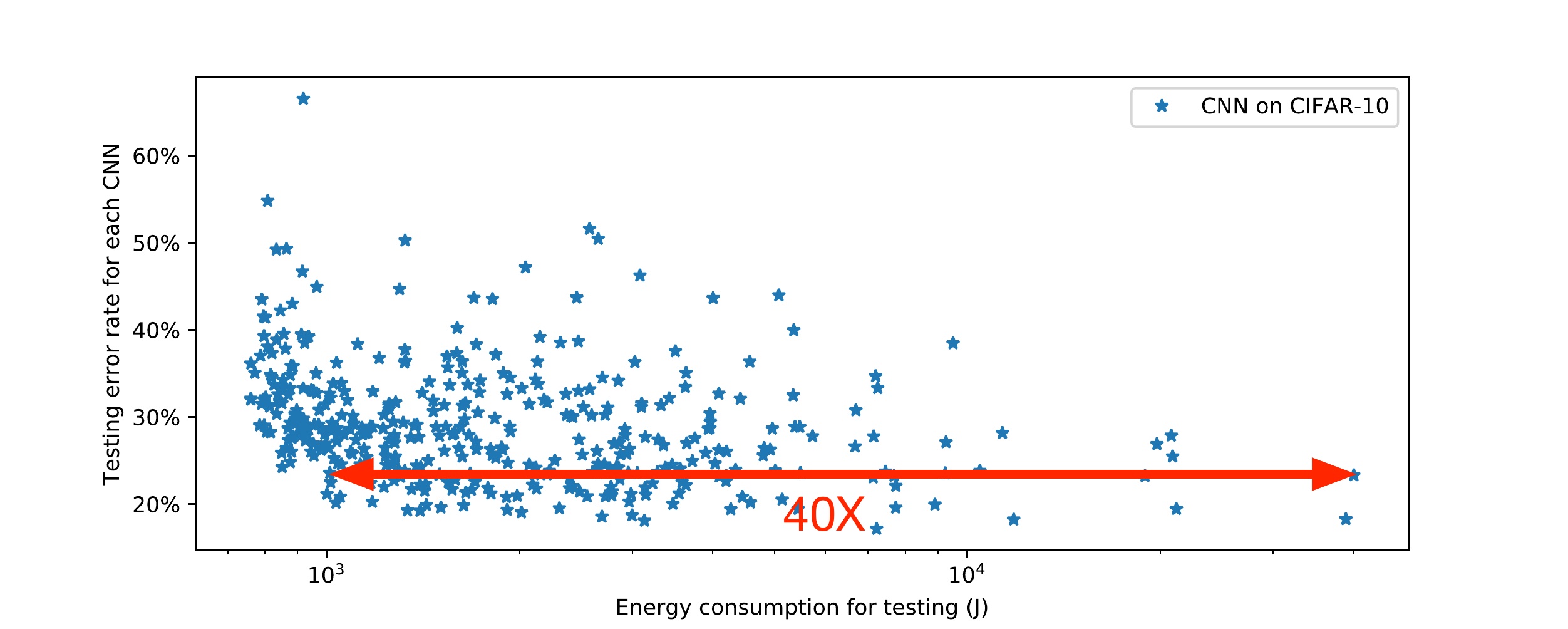}	
	\caption{Testing error vs. energy consumption for various CNN architectures on CIFAR-10 running with TensorFlow on Nvidia Titan X GPU. Each point represents one randomly-sampled CNN architecture for making inferences on CIFAR-10. For the architectures that achieve similar error rates (around 20\%) during test phase, the energy consumption can vary significantly by more than 40$\times$.}
	\label{fig:motivation}
\end{figure}

\added{In this paper, we focus on the \emph{testing} or \emph{service} phase since, CNNs are typically deployed to provide services (\textit{e.g.}, image recognition) that can potentially be invoked billions of times on millions of devices using the same architecture. Therefore, testing runtime and energy are critical to both users and cloud service providers. In contrast, training a CNN is usually done once. 
Orthogonal to many methods utilizing hardware characteristics to reduce energy consumptions, CNN architecture optimization in the design phase is significant. In fact, given the same performance level (\emph{e.g.}, the prediction accuracy in image recognition task), there are usually many CNNs with different energy consumptions. Figure~\ref{fig:motivation} shows the relationship between model testing errors and energy consumption for a variety of CNN architectures. We observe that several architectures can achieve a similar accuracy level. However, the energy consumption drastically differs among these architectures, with the difference as large as 40$\times$ in several cases. Therefore, seeking for energy-efficient CNN architecture without compromising performance seems intriguing, especially for large-scale deployment.}

\added{To identify energy-efficient CNNs, the use of accurate runtime, power, and energy models is of crucial importance. The reason for this is twofold. First, commonly used metrics characterizing CNN complexity (\emph{e.g.}, total FLOPs}\footnote{FLOP stands for ``floating point operation''.}\added{) are too crude to predict energy consumption for real platforms. Energy consumption depends not only on the CNN architecture, but also on the software/hardware platform. In addition, it is also hard to predict the corresponding runtime and power. Second, traditional profiling methods have limited effectiveness in identifying energy-efficient CNNs, due to several reasons: 1) These methods tend to be inefficient when the search space is large (\emph{e.g.}, more than 50 architectures); 2) They fail to quantitatively capture how changes in the CNN architectures affect runtime, power, and energy. Such results are critical in many automatic neural architecture search algorithms \cite{zoph2016neural}; 3) Typical CNNs are inconvenient or infeasible to profile if the service platform is different than the training platform. Therefore, it is imperative to train models for power, runtime, energy consumption of CNNs. Such models would significantly help Machine Learning practitioners and developers to design accurate, fast, and energy-efficient CNNs, especially in the design space of mobile or embedded platforms, where power- and energy-related issues are further exacerbated.}

\begin{figure}[ht!] 
	\centering
	\small
	\includegraphics[width=0.99\linewidth]{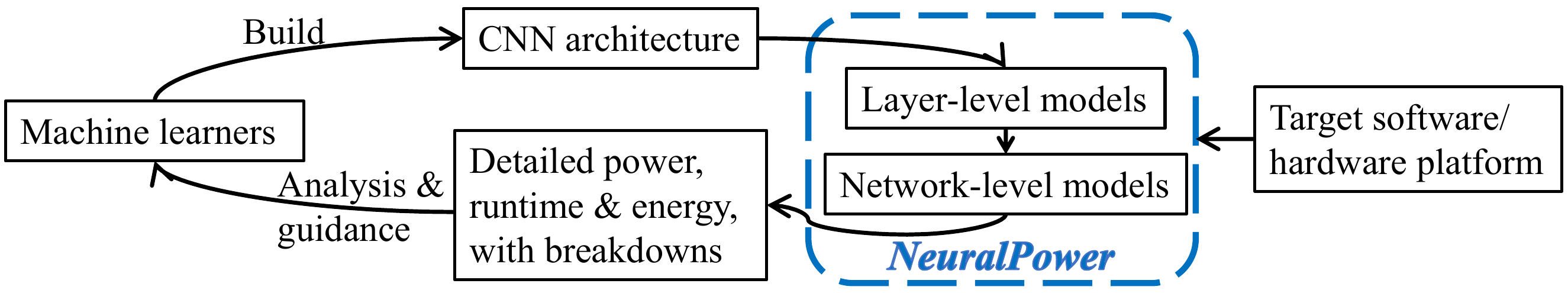}	
	\caption{\NP quickly predicts the power, runtime, and energy consumption of a CNN architecture during service phase. Therefore, \NP provides the machine learners with analysis and guidance when searching for energy-efficient CNN architectures on given software/hardware platforms.}
	\label{fig:neural_power}
\end{figure}

In this paper, we develop a predictive framework for power, runtime, and energy of CNNs during the testing phase, namely \NP, \emph{without} actually running (or implementing) these CNNs on a target platform. The framework is shown in Figure \ref{fig:neural_power}. That is, given (a) a CNN architecture of interest and (b) the target platform where the CNN model will be deployed, \NP can directly predict the power, runtime, and energy consumption of the network in service/deployment phase.
This paper brings the following contributions:
\begin{itemize}
\item To the best of our knowledge, our proposed learning-based polynomial regression approach, namely \NP, is the first framework for predicting the \textit{power consumption} of CNNs running on GPUs, with an average accuracy of 88.34\%.
\item \NP can also predict \textit{runtime} of CNNs, which outperforms state-of-the-art analytical models, by achieving an improvement in accuracy up to $68.5\%$ compared to the best previously published work.

\item \NP uses power and runtime predictions to predict the \emph{energy consumption} of CNNs running on GPUs, with an average accuracy of 97.21\%. 

\item In addition to total runtime and average power, \NP also provides the detailed breakdown of runtime and power across different components (at each layer) of the whole network, thereby helping machine learners to identify efficiently the runtime, power, or energy bottlenecks.

\item We propose the \emph{Energy-Precision Ratio} metric for guiding machine learners to trade-off between CNN classification accuracy and energy efficiency.
\end{itemize}

The rest of this paper is organized as follows. Section \ref{sec:background} introduces the background knowledge and related work. Section \ref{sec:models} provides the details of the proposed \NP: layer-level and whole-network models, and the process (configurations and platforms) for collecting the dataset. Section \ref{sec:results} shows the experimental results on power, runtime, and energy predictions. Finally, Section \ref{sec:conclusion} concludes the paper.

\section{Background and Related Work}
\label{sec:background}
Prior art, \emph{e.g.}, by \cite{han2015learning}, has identified the runtime overhead and power consumption of CNNs execution to be a significant concern when designing accurate, yet power- and energy-efficient CNN models. These design constraints become increasingly challenging to address, especially in the context of two emerging design paradigms, \emph{(i)} the design of larger, power-consuming CNN architectures, and \emph{(ii)} the design of energy-aware CNNs for mobile platforms. Hence, a significant body of state-of-the-art approaches aim to tackle such runtime and power overhead, by accelerating the execution of CNNs and/or by reducing their power-energy consumption.

\textbf{Runtime-efficient CNNs}: To enable CNN architectures that execute faster, existing work has investigated both hardware- and software-based methodologies. On one hand, with respect to hardware, several hardware platforms have been explored as means to accelerate the CNN execution. Examples of these platforms include FPGA-based methodologies by \cite{zhang2015fpgas}, ASIC-like designs by~\cite{zhang2016asic}. On the other hand, with respect to software-based acceleration, several libraries, such as cuDNN and Intel MKL, have been used in various deep learning frameworks to enable fast execution of CNNs, since these libraries provide specially-optimized primitives in CNNs to improve their runtime. Among the different hardware and software platforms, GPU-based frameworks are widely adopted both in academia and industry, thanks to the good trade-off between platform flexibility and performance that they provide. To this end, in this work, we aim to model the runtime and power characteristic of CNNs executing on state-of-the-art GPU platforms.

\textbf{Power- and energy-aware CNNs}: Recent work has focused on limiting the energy and power consumption of CNNs. 
Several approaches investigate the problem in the context of hyper-parameter optimization. For instance, \cite{rouhani2016delight} have proposed an automated customization methodology that adaptively conforms the CNN architectures to the underlying hardware characteristics, while minimally affecting the inference accuracy. \cite{smithson2016design} have used a pair of trainer-trainee CNNs to incorporate accuracy and cost into the design of neural networks. Beyond these methods which are based on hyper-parameter optimization, another group of novel approaches focuses solely on the energy-efficient CNN implementation assuming a given network architecture. These approaches include techniques that draw ideas from energy-aware computer system design, such as the methodologies by~\cite{han2015learning} and ~\cite{courbariaux2016binarized}. 

\textbf{Modeling of metrics characterizing CNNs}: While all the aforementioned approaches motivate hardware-related constraints as limiting factors towards enabling efficient CNN architecture, to the best of our knowledge there is no comprehensive methodology that models the runtime, power, and eventually the energy of CNN architectures. That is, prior work either relies on proxies of memory consumption or runtime expressed as simplistic counts of the network weights (\emph{e.g.}, as done by \cite{rouhani2016going}), or extrapolates power-energy values from energy-per-operation tables (as assumed by~\cite{han2015learning}). Consequently, existing modeling assumptions are either overly simplifying or they rely on outdated technology nodes. Our work successfully addresses these limitations, by proposing power, energy, and runtime models that are validated against state-of-the-art GPUs and Deep Learning software tools.

A work that shares similar insight with our methodology is the \textit{Paleo} framework proposed by~\cite{qi2016paleo}. In their approach, the authors present an analytical method to determine the runtime of CNNs executing on various platforms. However, their model cannot be flexibly used across different platforms with different optimization libraries, without detailed knowledge of them. More importantly, their approach cannot predict power and energy consumption. \deleted{As the best published methodology to date, we use this work as our baseline in our results. We show that our proposed methodology \emph{advances the state-of-the-art} by achieving better prediction accuracy of the runtime of CNNs on various platforms, and by proposing an accurate predictive model for power and energy.}

\section{Methodology: Power and Runtime Modeling}
\label{sec:models}
In this section, we introduce our hierarchical power and runtime model framework -\NP- and the data collection process. \NP is based on the following key insight: despite the huge amount of different CNN variations that have been used in several applications, all these CNN architectures consist of basic underlying building blocks/primitives which exhibit similar execution characteristics per type of layer. To this end, \NP first models the power and runtime of the key layers that are commonly used in a CNN. Then, \NP uses these models to predict the performance and runtime of the entire network. 

\subsection{Layer-Level Power and Runtime Modeling}
\label{subsec:layer-wise model}
The first part of \NP is layer-level power and runtime models. We construct these models for each type of layer for both runtime and power. More specifically, we select to model three types of layers, namely the \textit{convolutional}, the \textit{fully connected}, and the \textit{pooling} layer, since these layers carry the main computation load during CNN execution -- as also motivated by \cite{qi2016paleo}. Nevertheless, unlike prior work, our goal is to make our model flexible for various combinations of software/hardware platforms without knowing the details of these platforms. 

To this end, we propose a learning-based \textit{polynomial regression model} to learn the coefficients for different layers, and we assess the accuracy of our approach against power and runtime measurements on different commercial GPUs and Deep Learning software tools. 
There are three major reasons for this choice. \emph{First}, in terms of model accuracy, polynomial models provide more flexibility and low prediction error when modeling both power and runtime. The \emph{second} reason is the interpretability: runtime and power have clear physical correlation with the layer's configuration parameters (\emph{e.g.}, batch size, kernel size, \emph{etc.}). That is, the features of the model can provide an intuitive encapsulation of how the layer parameters affect the runtime and power. The \emph{third} reason is the available amount of sampling data points. 
Polynomial models allow for adjustable model complexity by tuning the degree of the polynomial, ranging from linear model to polynomials of high degree, whereas a formulation with larger model capacity may be prone to overfitting. To perform model selection, we apply ten-fold cross-validation and we use Lasso to decrease the total number of polynomial terms. \added{The detailed model selection process will be discussed in Section \ref{subsec:results:layer-wise}.}

\textbf{Layer-level runtime model}: The runtime $\hat{T}$ of a layer can be expressed as: 
\begin{align}
  \label{eq:polynomial_runtime}
  \hat{T}(\bx_T)  =  & \sum _{j}  c_j \cdot  \prod_{i = 1}^{D_T} {x}_i^{q_{ij}} + \sum_s c^\prime_s \mathcal{F}_s(\bx_T)\\
\text{where  } & \bx_T \in \mathbb{R}^{D_T}; \ q_{ij} \in \mathbb{N}; \  \forall j, \ \sum_{i = 1}^{D_T} q_{ij} \leq K_T. \nonumber
\end{align}
The model consists of two components. The first component corresponds to the regular degree-$K_T$ polynomial terms which are a function of the features in input vector $\bx_T \in \mathbb{R}^{D_T}$. $x_i$ is the $i$-th component of $\bx_T$. $q_{ij}$ is the exponent for $x_i$ in the $j$-th polynomial term, and $c_j$ is the coefficient to learn. This feature vector of dimension $D_T$ includes layer configuration hyper-parameters, such as the batch size, the input size, and the output size. For different types of layers, the dimension $D_T$ is expected to vary. For convolutional layers, for example, the input vector includes the kernel shape, the stride size, and the padding size, whereas such features are not relevant to the formulation/configuration of a fully-connected layer.

The second component corresponds to special polynomial terms $\mathcal{F}$, which encapsulate physical operations related to each layer (\emph{e.g.}, the total number of memory accesses and the total number of floating point operations). 
The number of the special terms differs from one layer type to another. For the convolutional layer, for example, the special polynomial terms include the memory access count for input tensor, output tensor, kernel tensor, and the total number of floating point operations for the all convolution computations. Finally, $c^\prime_s$ is the coefficient of the $s$-th special term to learn.

Based on this formulation, it is important to note that not all input parameters are positively correlated with the runtime. For example, if the stride size increases, the total runtime will decrease since the total number of convolutional operations will decrease. This observation motivates further the use of a polynomial formulation, since it can capture such trends (unlike a posynomial model, for instance). \deleted{Please also note that we use Equation~\ref{eq:polynomial_runtime} to predict the runtime of a CNN during \emph{testing}. This is because (as already motivated in Section \ref{sec:introduction}) the runtime of CNNs at service time is a design metric of crucial importance, especially when designing for energy-constrained mobile platforms or for more complex CNN configurations. However, we can flexibly use a similar formulation to model the runtime during training (\emph{i.e.}, by simply incorporating a model for back-propagation runtime), whose investigation we leave for future work.}

\textbf{Layer-level power model}: To predict the power consumption $\hat{P}$ for each layer type during \emph{testing}, we follow a similar polynomial-based approach:
\begin{align}
  \label{eq:polynomial_power}
  \hat{P}(\bx_P)  =  & \sum _{j}  z_j \cdot  \prod_{i = 1}^{D_P} {x}_i^{m_{ij}} + \sum_k z^\prime_k \mathcal{F}_k (\bx_P)\\
\text{where  } & \bx_P \in \mathbb{R}^{D_P}; \ m_{ij} \in \mathbb{N}; \  \forall j, \ \sum_{i = 1}^{D_P} m_{ij} \leq K_P. \nonumber
\end{align}
where the regular polynomial terms have degree $K_P$ and they are a function of the input vector $\bx_P \in \mathbb{R}^{D_P}$. $m_{ij}$ is the exponent for $x_i$ of the $j$-th polynomial term, and $z_j$ is the coefficient to learn. In the second sum, $z^\prime_k$ is the coefficient of the $k$-th special term to learn.

Power consumption however has a non-trivial correlation with the input parameters. More specifically, as a metric, power consumption has inherent limits, \emph{i.e.}, it can only take a range of possible values constrained through the power budget. That is, when the computing load increases, power does not increase in a linear fashion. To capture this trend, we select an extended feature vector $\bx_P \in \mathbb{R}^{D_P}$ for our power model, where we include the logarithmic terms of the features used for runtime (\emph{e.g.}, batch size, input size, output size, \emph{etc.}). As expected, the dimension $D_P$ is twice the size of the input feature dimension $D_T$. A logarithmic scale in our features vector can successfully reflect such a trend, as supported by our experimental results in Section \ref{sec:results}.

\subsection{Network-Level Power, Runtime, and Energy Modeling}
\label{subsec:whole_network}
\added{We discuss the network-level models for \NP.} For the majority of CNN architectures readily available in a Deep Learning models ``zoo'' (as the one compiled by ~\cite{jia2014caffe}), the whole structure consists of and can be divided into several layers in series. Consequently, using our predictions for power and runtime as building blocks, we extend our predictive models to capture the runtime, the power, and eventually the energy, of the entire architecture at the \emph{network level}. 

\textbf{Network-level runtime model}: Given a network with $N$ layers connected in series, the predicted total runtime can be written as the sum of the predicted runtime $\hat{T}_n$ of each layer $n$:
\begin{equation}
\label{eq:runtime}
\hat{T}_{total} = \sum_{n = 1}^N \hat{T}_n
\end{equation}
\textbf{Network-level power model}: Unlike the summation for total runtime, the average power can be obtained using both per layer runtime and power. More specifically, we can represent the average power $\hat{P}_{avg}$ of a CNN as:
\begin{equation}
\label{eq:power}
\hat{P}_{avg} = \frac{\sum_{n = 1}^N \hat{P}_n \cdot \hat{T}_n}{\sum_{n = 1}^N \hat{T}_n}
\end{equation}
\textbf{Network-level energy model}: From here, it is easy to derive our model for the total energy consumption $\hat{E}_{total}$ of an entire network configuration:
\begin{equation}
\label{eq:energy}
\hat{E}_{total} =  \hat{T}_{total} \cdot \hat{P}_{avg} = \sum_{n = 1}^N \hat{P}_n \cdot \hat{T}_n
\end{equation}
which is basically the scalar product of the layer-wise power and runtime vectors, or the sum of energy consumption for all layers in the model.

\subsection{Dataset Collection}
\label{subsec:setup}

\textbf{Experiment setup}: 
The main modeling and evaluation steps are performed on the platform described in Table \ref{tab:Target architecture}. To exclude the impact of voltage/frequency changing on the power and runtime data we collected, we keep the GPU in a fixed state and CUDA libraries ready to use by enabling the persistence mode. We use \texttt{nvidia-smi} to collect the instantaneous power per 1 ms for the entire measuring period. 
Please note that while this experimental setup constitutes our configuration basis for investigating the proposed modeling methodologies, in Section~\ref{subsec:models on other platforms} we present results of our approach on other GPU platforms, such as Nvidia GTX 1070, and Deep Learning tools, such as Caffe by~\cite{jia2014caffe}.

\begin{table}[ht]	
	\centering
	\vspace{-6pt}
	\caption{Target platform}
	\label{tab:Target architecture}
		\small
		\begin{tabular}
			{l|l} \toprule
            \textbf{CPU / Main memory} & Intel Core-i7 5820K / 32GB \\ \hline
			\textbf{GPU} & Nvidia GeForce GTX Titan X (12GB DDR5) \\ \hline
			\textbf{GPU max / idle power} & 250W / 15W \\ \hline
			\textbf{Deep learning platform} & TensorFlow 1.0 on Ubuntu 14 \\ \hline
			\textbf{Power meter} & NVIDIA System Management Interface\\ \bottomrule
		\end{tabular}	 
\end{table}

\textbf{CNN architectures investigated}: To comprehensively assess the effectiveness of our modeling methodology, we consider several CNN models. Our analysis includes state-of-the-art configurations, such as the AlexNet by~\cite{krizhevsky2014one}, VGG-16 \& VGG-19 by~\cite{simonyan2014very}, R-CNN by~\cite{ren2015faster}, NIN network by~\cite{lin2013network}, CaffeNet by~\cite{jia2014caffe}, GoogleNet by~\cite{szegedy2015going}, and Overfeat by~\cite{sermanet2013overfeat}. We also consider different flavors of smaller networks such as vanilla CNNs used on the MNIST by~\cite{lecun1998gradient} and CIFAR10-6conv~\cite{courbariaux2015binaryconnect} on CIFAR-10. This way, we can cover a wider spectrum of CNN applications. 

\textbf{Data collection for power/runtime models}: To train the layer-level predictive models, we collect data points by profiling power and runtime from all layers of all the considered CNN architectures in the training set. We separate the training data points into groups based on their layer types. \added{In this paper, the training data include 858 convolution layer samples, 216 pooling layer samples, and 116 fully connected layer samples. The statistics can change if one needs any form of customization. }For testing, we apply our learned models on the network in the testing set, and compare our predicted results against the actual results profiled on the same platform, including both layer-level evaluation and network-level evaluation.

\section{Experimental Results} 
\label{sec:results}
In this section, we assess our proposed \NP in terms of power, runtime, and energy prediction accuracy at both layer level and network level. Since the models for runtime and power are slightly different from one to another, we discuss them separately in each case. We then propose a metric called \emph{Energy-Precision Ratio} to guide machine learners toward energy-efficient CNNs. In the end of this section, we validate our framework on other hardware and software platforms.

\subsection{Layer-Level Model Evaluation}
\label{subsec:results:layer-wise}
\subsubsection{Model selection}
\begin{figure}[htbp] 
	\centering
	\small
	\includegraphics[width=0.99\linewidth]{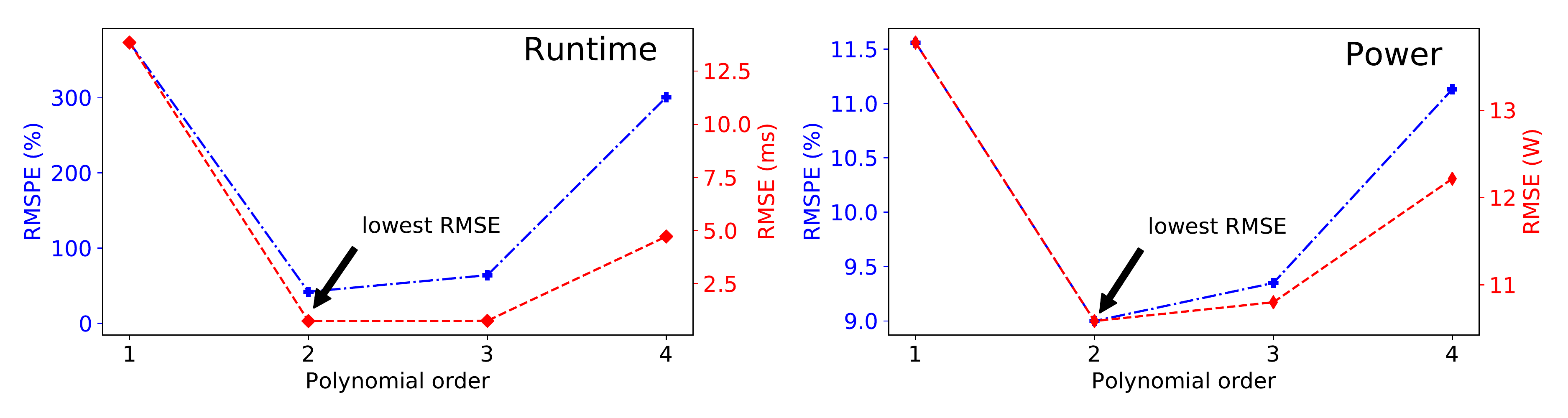}	
	\vspace{-15pt}
	\caption{Comparison of best-performance model with respect to each polynomial order for the fully-connected layers. In this example, a polynomial order of two is chosen since it achieves the best Root-Mean-Square-Error (RMSE) for both runtime and power modeling. At the same time, it also has the lowest Root-Mean-Square-Percentage-Error (RMSPE).}
	\label{fig:model_selection}
\end{figure}

\added{To begin with model evaluation, we first illustrate how model selection has been employed in \NP. In general, \NP changes the order of the polynomial (\emph{e.g.}, $D_T$ in Equation \ref{eq:polynomial_runtime}) to expand/shrink the size of feature space. \NP applies Lasso to select the best model for each polynomial model. Finally, \NP selects the final model with the lowest cross-validation Root-Mean-Square-Error (RMSE), which is shown in Figure \ref{fig:model_selection}.}

\subsubsection{Runtime Models}
Applying the model selection process, we achieve a polynomial model for each layer type in a CNN. The evaluation of our models is shown in Table~\ref{tab:perf_model}, where we report the Root-Mean-Square-Error (RMSE) and the relative Root-Mean-Square-Percentage-Error (RMSPE) of our runtime predictions for each one of the considered layers. Since we used Lasso in our model selection process, we also report the model size (\emph{i.e.}, the number of terms in the polynomial) per layer. More importantly, we compare against the state-of-the-art analytical method proposed by \cite{qi2016paleo}, namely Paleo. To enable a comparison here and for the remainder of section, we executed the Paleo code\footnote{Paleo is publicly available online -- github repository: \url{https://github.com/TalwalkarLab/paleo}} on the considered CNNs. We can easily observe that our predictions based on the layer-level models clearly outperform the best published model to date, yielding an \emph{improvement in accuracy} up to $68.5\%$. 

\begin{table}[ht]	
	\centering
	\caption{Comparison of runtime models for common CNN layers -- Our proposed runtime model consistently outperforms the state-of-the-art runtime model in both root-mean-square-error (RMSE) and the Root-Mean-Square-Percentage-Error (RMSPE).} 
	\label{tab:perf_model}
		\small
		\begin{tabular}
			{l|ccc|cc} \toprule
			\multirow{2}{*}{Layer type } & \multicolumn{3}{c}{\textbf{\NP}} & \multicolumn{2}{|c}{Paleo~\cite{qi2016paleo}} \\ \cline{2-4}\cline{5-6}
            & Model size & RMSPE & RMSE (ms) & RMSPE & RMSE  (ms)\\ \hline
			Convolutional & 60 & 39.97\%	& 1.019  & 58.29\% & 4.304\\ 
			Fully-connected & 17 & 41.92\% & 0.7474 & 73.76\% & 0.8265\\ 
			Pooling & 31 & 11.41\% & 0.0686  & 79.91\% &1.763 \\ \bottomrule
		\end{tabular}	 
\end{table}

\textbf{Convolutional layer}: The convolution layer is among the most time- and power-consuming components of a CNN. To model this layer, we use a polynomial model of degree three. We select a features vector consisting of the batch size, the input tensor size, the kernel size, the stride size, the padding size, and the output tensor size. In terms of the special terms defined in Equation~\ref{eq:polynomial_runtime}, we use terms that represent the total computation operations and memory accesses count. \textbf{Fully-connected layer}: We employ a regression model with degree of two, and as features of the model we include the batch size, the input tensor size, and the output tensor size. It is worth noting that in terms of software implementation, there are two common ways to implement the fully-connected layer, either based on default matrix multiplication, or based on a convolutional-like implementation (\emph{i.e.}, by keeping the kernel size exactly same as the input image patch size). Upon profiling, we notice that both cases have a tensor-reshaping stage when accepting intermediate results from a previous convolutional layer, so we treat them interchangeably under a single formulation. \textbf{Pooling layer}: The pooling layer usually follows a convolution layer to reduce the complexity of the model. As basic model features we select the input tensor size, the stride size, the kernel size, and the output tensor size. Using Lasso and cross-validation we find that a polynomial of degree three provides the best accuracy.

\subsubsection{Power Models}
As mentioned in Section \ref{subsec:layer-wise model}, we use the logarithmic terms of the original features (\emph{e.g.}, batch size, kernel size, \emph{etc.}) as additional features for the polynomial model since this significantly improves the model accuracy. This modeling choice is well suited for the nature of power consumption which does not scale linearly; more precisely, the rate of the increase in power goes down as the model complexity increases,  especially when the power values get closer to the power budget limit. For instance, in our setup, the Titan X GPU platform has a maximum power of 250W. We find that a polynomial order of two achieves the best cross validation error for all three layer types under consideration.

To the best of our knowledge, there is no prior work on power prediction at the layer level to compare against. We therefore compare our methods directly with the actual power values collected from TensorFlow, as shown in Table~\ref{tab:power_model}. Once again, we observe that our proposed model formulation achieves error always less that $9\%$ for all three layers. The slight increase in the model size compared to the runtime model is to be expected, given the inclusion of the logarithmic feature terms, alongside special terms that include memory accesses and operations count. We can observe, though, that the model is able to capture the trends of power consumption trained across layer sizes and types.

\begin{table}[ht]	
	\centering
	\caption{Power model for common CNN layers}
	\label{tab:power_model}
			\small
		\begin{tabular}
			{lccc} \toprule
			\multirow{2}{*}{Layer type } & \multicolumn{3}{c}{\textbf{\NP}}  \\
            \cline{2-4}
            & Model size & RMSPE & RMSE (W) \\ \hline
			Convolutional & 75 & 7.35\%	& 10.9172 \\ 
			Fully-connected & 15 & 9.00\% & 10.5868 \\ 
			Pooling & 30 & 6.16\% & 6.8618\\ \bottomrule
		\end{tabular}	 
\end{table}

\subsection{Network-level Modeling Evaluation}
With the results from layer-level models, we can model the runtime, power, and energy for the whole network based on the network-level model (Section~\ref{subsec:whole_network}) in \NP. To enable a comprehensive evaluation, we assess \NP on several state-of-the-art CNNs, and we compare against the actual runtime, power, and energy values of each network. For this purpose, and as discussed in Section~\ref{subsec:setup}, we leave out a set of networks to be used only for testing, namely the VGG-16, NIN, CIFAR10-6conv, AlexNet, and Overfeat networks.

\subsubsection{Runtime evaluation}
\textbf{Enabling network runtime profiling}: Prior to assessing the predictions on the networks as a whole, we show the effectiveness of \NP as a useful aid for CNN architecture benchmarking and per-layer profiling. Enabling such breakdown analysis is significant for machine learning practitioners, since it allows to identify the bottlenecks across components of a CNN. \deleted{Hence, we show first that \NP can be used for the accurate profiling of different components of a network of interest, outperforming the best published model to date in accuracy. }

\begin{figure}[htbp] 
	\centering
	\small
	\includegraphics[width=0.99\linewidth]{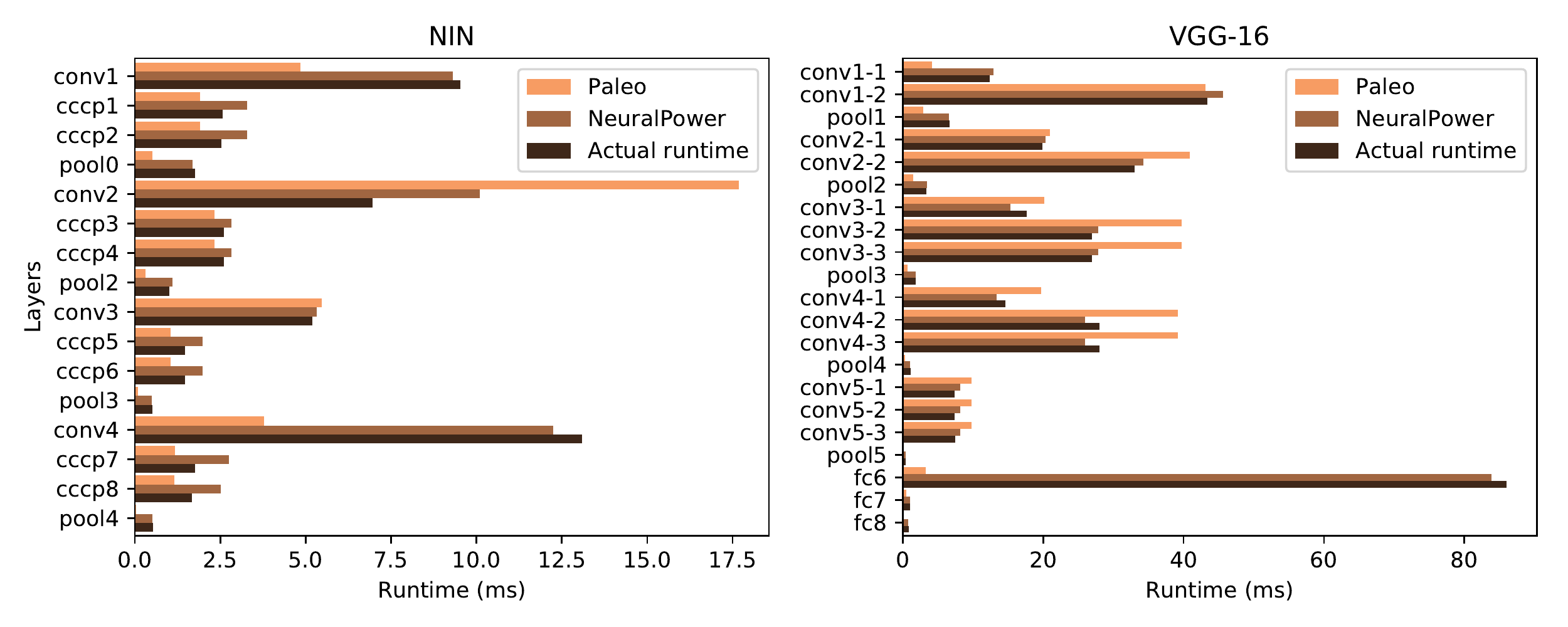}	
	\vspace{-15pt}
	\caption{Comparison of runtime prediction for each layer in NIN and VGG-16: Our models provide accurate runtime breakdown of both network, while Paleo cannot. Our model captures the execution-bottleneck layers (\emph{i.e.}, \emph{conv4} in NIN, and \emph{fc6} in VGG-16) while Paleo mispredicts both.}
	\label{fig:runtime}
\end{figure}

For runtime, we use state-of-the-art analytical model Paleo as the baseline. In Figure~\ref{fig:runtime}, we compare runtime prediction models from \NP and the baseline against actual runtime values of each layer in the NIN and VGG-16 networks. From Figure \ref{fig:runtime}, we can clearly see that our model outperforms the Paleo model for most layers in accuracy. For the NIN, our model clearly captures that \textit{conv4} is the dominant (most time-consuming) layer across the whole network. However, Paleo erroneously identifies \textit{conv2} as the dominant layer. For the VGG-16 network, we can clearly see that Paleo predicts the runtime of the first fully-connected layer \textit{fc6} as 3.30 ms, with a percentage prediction error as high as -96.16\%. In contrast, our prediction exhibits an error as low as -2.53\%. Since layer \textit{fc6} is the dominant layer throughout the network, it is critical to make a correct prediction on this layer. 

From the above, we can conclude that our proposed methodology generally has a better accuracy in predicting the runtime for each layer in a complete CNN, especially for the layers with larger runtime values. Therefore, our accurate runtime predictions, when employed for profiling each layer at the network level, can help the machine learners and practitioners quickly identify the real bottlenecks with respect to runtime for a given CNN.

\textbf{Network-level runtime evaluation}: Having demonstrated the effectiveness of our methodology at the layer level, we proceed to assess the accuracy of the network-level runtime prediction $\hat{T}_{total}$ (Equation~\ref{eq:runtime}). It is worth observing that in Equation~\ref{eq:runtime} there are two sources of potential error. First, error could result from mispredicting the runtime values $\hat{T}_{n}$ per layer $n$. However, even if these predictions are correct, another source of error could come from the formulation in Equation~\ref{eq:runtime}, where we assume that the sum of the runtime values of all the layers in a CNN provides a good estimate of the total runtime. Hence, to achieve a comprehensive evaluation of our modeling choices in terms of both the regression formulation and the summation in Equation~\ref{eq:runtime}, we need to address both these concerns. 

To this end, we compare our runtime prediction $\hat{T}_{total}$ against two metrics. \emph{First}, we compare against the actual overall runtime value of a network, notated as $T_{total}$. \emph{Second}, we consider another metric defined as the sum of the actual runtime values $T_n$ (and not the predictions) of each layer $n$:
\begin{equation}
\label{eq:runtime_comparison}
\mathbb{T}_{total} = \sum_{n = 1}^N T_n
\end{equation}
Intuitively, a prediction value $\hat{T}_{total}$ close to both the $\mathbb{T}_{total}$ value and the actual runtime $T_{total}$ would not only show that our model has good network-level prediction, but that also that our underlying modeling assumptions hold. 

We summarize the results across five different networks in Table \ref{tab:whole_model_runtime}. More specifically, we show the networks' actual total runtime values ($T_{total}$), the runtime $\mathbb{T}_{total}$ values, our predictions $\hat{T}_{total}$, and the predictions from Paleo (the baseline). Based on the Table, we can draw two key observations. First, we can clearly see that our model always outperforms Paleo, with runtime predictions always within $24\%$ from the actual runtime values. Unlike our model, prior art could underestimate the overall runtime up to $42\%$. Second, as expected, we see that summing the true runtime values per layer does indeed approximate the total runtime, hence confirming our assumption in Equation~\ref{eq:runtime}. 

\vspace*{-6pt}
\begin{table}[ht]	
	\centering
	\caption{Performance model comparison for the whole network. We can easily observe that our model always provides more accurate predictions of the total CNN runtime compared to the best published model to date (Paleo). We assess the effectiveness of our model in five different state-of-the-art CNN architectures.}
	\label{tab:whole_model_runtime}
		\small
		\begin{tabular}
			{c|c|c|c||c} \toprule
			CNN  & \cite{qi2016paleo} & \textbf{\NP}     &    Sum of per-layer actual    & Actual runtime  \\ 
			 name  & Paleo (ms)       & $\hat{T}_{total}$ (ms) & runtime $\mathbb{T}_{total}$ (ms) & $T_{total}$ (ms)  \\ \hline
			VGG-16 & 345.83 & 373.82 & 375.20 &  368.42 \\ 
			AlexNet & 33.16  & 43.41 & 42.19 & 39.02 \\ 
			NIN & 45.68  & 62.62 & 55.83 & 50.66 \\ 
			Overfeat & 114.71  & 195.21 & 200.75 & 197.99 \\ 
			CIFAR10-6conv & 28.75  & 51.13 & 53.24 & 50.09 \\ \bottomrule
		\end{tabular}	 
\end{table}

\subsubsection{Power evaluation}

\begin{figure}[htbp] 
	\centering
	\small
	\includegraphics[width=0.99\linewidth]{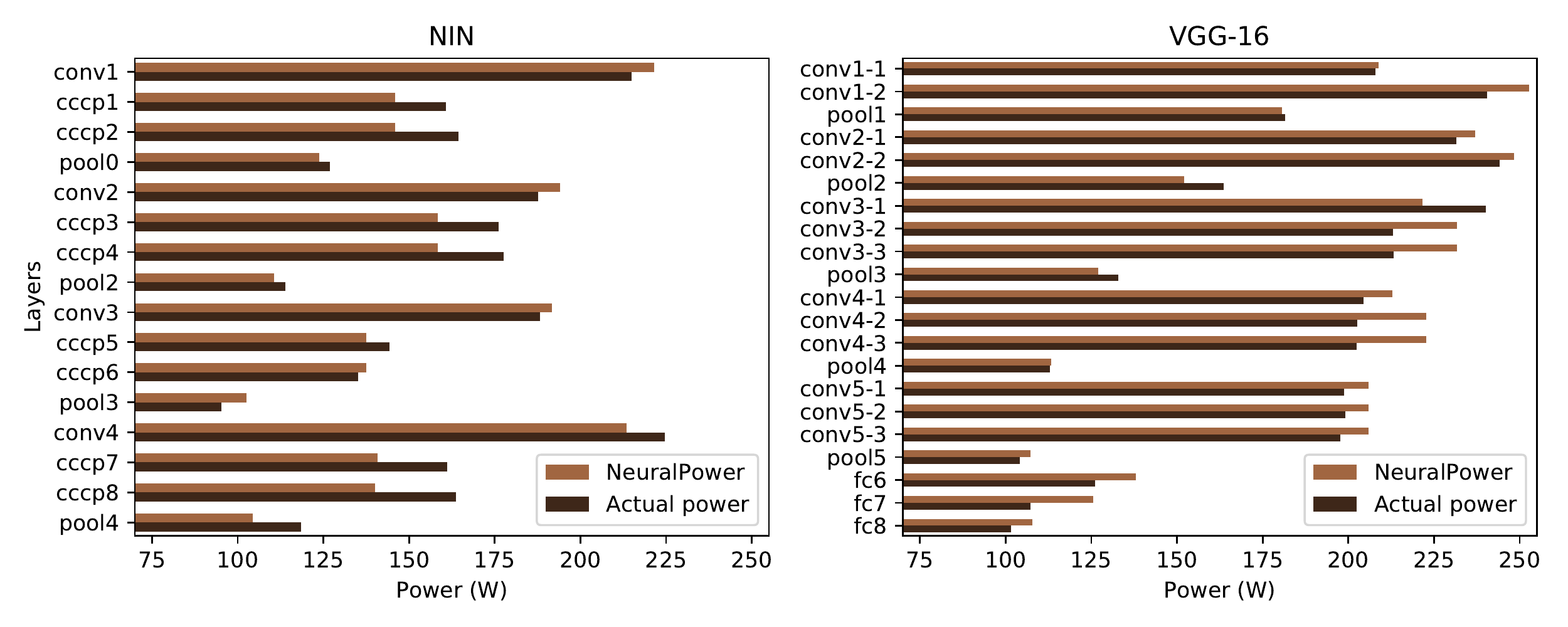}	
	\vspace{-15pt}
	\caption{Comparison of power prediction for each layer in NIN and VGG-16.}
	\label{fig:power}
\end{figure}

\textbf{Enabling power network profiling}: We present a similar evaluation methodology to assess our model for network-level power predictions. We first use our methodology to enable a per-layer benchmarking of the power consumption. Figure \ref{fig:power} shows the comparison of our power predictions and the actual power values for each layer in the NIN and the VGG-16 networks. 
We can see that convolutional layers dominate in terms of power consumption, while pooling layers and fully connected layers contribute relatively less. We can also observe that the convolutional layer exhibits the largest variance with respect to power, with power values ranging  from 85.80W up to 246.34W. 

Another key observation is related to the fully-connected layers of the VGG-16 network. From Figure \ref{fig:runtime}, we know layer \textit{fc6} takes the longest time to run. Nonetheless, we can see in Figure \ref{fig:power} that its power consumption is relatively small. Therefore, the energy consumption related of layer \textit{fc6} will have a smaller contribution to the total energy consumption of the network relatively to its runtime. It is therefore evident that using only the runtime as a proxy proportional to the energy consumption of CNNs could mislead the machine learners to erroneous assumptions. This observation highlights that power plays also a key role towards representative benchmarking of CNNs, hence illustrating further the significance of accurate power predictions enabled from our approach.

\begin{table}[ht]	
	\centering
	\vspace{-12pt}
	\caption{Evaluating our power predictions for state-of-the-art CNN architectures.}
	\label{tab:whole_model_power}
    	\small
		\begin{tabular}
			{c|c|c||c} \toprule
			CNN  &  \textbf{\NP}     &    Sum of per-layer actual    & Actual power  \\ 
			name  & $\hat{P}_{total}$ (W) & power $\mathbb{P}_{total}$ (W) & $P_{avg}$ (W)  \\ \hline
			VGG-16 & 206.88 & 195.76 & 204.80 \\ 
			AlexNet & 174.25 & 169.08 & 194.62 \\ 
			NIN & 179.98 & 187.99 & 226.34 \\ 
			Overfeat  & 172.20 & 168.40 & 172.30 \\ 
			CIFAR10-6conv  & 165.33 & 167.86 & 188.34 \\ \bottomrule
		\end{tabular}	 
\end{table}

\textbf{Network-level power evaluation}: As discussed in the runtime evaluation as well, we assess both our predictive model accuracy and the underlying assumptions in our formulation. In terms of average power consumption, we need to confirm that the formulation selected in Equation~\ref{eq:power} is indeed representative. To this end, besides the comparison against the actual average power of the network $P_{avg}$, we compare against the average value $\mathbb{P}_{avg}$, which can be computed by replacing our predictions $\hat{P}_n$ and $\hat{T}_n$ with with the actual per-layer runtime and power values:
\begin{equation}
\label{eq:power_comparison}
\mathbb{P}_{avg} = \frac{\sum_{n = 1}^N {P}_n \cdot {T}_n}{\sum_{n = 1}^N {T}_n}
\end{equation}
We evaluate our power value predictions for the same five state-of-the-art CNNs in Table \ref{tab:whole_model_power}. Compared to the actual power value, our prediction have an RMSPE of $11.66\%$. We observe that in two cases, AlexNet and NIN, our prediction has a larger error, \emph{i.e.}, of $10.47\%$ and $20.48\%$ respectively. This is to be expected, since our formulation for $\mathbb{P}_{avg}$ depends on runtime prediction as well, and as observed previously in Table~\ref{tab:whole_model_runtime}, we underestimate the runtime in both cases. 

\subsubsection{Energy evaluation}
Finally, we use Equation~\ref{eq:energy} to predict the total energy based on our model. To evaluate our modeling assumptions as well, we compute the energy value $\mathbb{E}_{total}$ based on the actual per-layer runtime and power values, defined as:
\begin{equation}
\label{eq:energy_comparison}
\mathbb{E}_{total} = \sum_{n = 1}^N {P}_n \cdot  {T}_n
\end{equation}
We present the results for the same five CNNs in Table~\ref{tab:whole_model_energy}. We observe that our approach enables good prediction, with an average RMSPE of $2.79\%$.

\begin{table}[ht]	
	\centering
	\caption{Evaluating our energy predictions for state-of-the-art CNN architectures.}
	\label{tab:whole_model_energy}
		\small
		\begin{tabular}
			{c|c|c||c} \toprule
			CNN  &  \textbf{\NP}   &    Sum of per-layer actual    & Actual energy  \\ 
			name  & $\hat{E}_{total}$ (J) & energy $\mathbb{E}_{total}$ (J) & $E_{total}$ (J)  \\ \hline
			VGG-16 	&77.312	& 73.446 & 75.452 \\ 
			AlexNet & 7.565 & 7.134 & 7.594 \\
			NIN &11.269 & 10.495	& 11.465 \\
			Overfeat &33.616 & 33.807 & 34.113 \\
			CIFAR10-6conv  & 8.938	& 8.453 & 9.433 \\ \bottomrule
		\end{tabular}	 
\end{table}

\subsection{Energy-Precision Ratio}
In this subsection, we propose a new metric, namely \textit{Energy-Precision Ratio} to be used as a guideline for machine learners towards accurate, yet energy efficient CNN architectures. We define the metric as:
\begin{equation}
M = Error^\alpha \cdot EPI
\end{equation}
where $Error$ is the data classification error of the CNN under consideration, and $EPI$ is the energy consumption per data item classified. Different values of the parameter $\alpha$ dictate how much importance is placed on the accuracy of the model, since a larger $\alpha$ places more weight on the CNN classification error. To illustrate how $\alpha$ affects the results, in Table \ref{tab:energy_ratio} we compute the $M$ score values for VGG-16, AlexNet, and NIN, all trained on ImageNet datasets (as $Error$ value we use their Top-5 error). We also use our predictive model for energy, and we compute the energy consumption per image classification. 

Intuitively, a lower $M$ value indicates a better trade-off between energy-efficiency and accuracy of a CNN architecture. For instance, we can see that while VGG-16 has the lowest error, this comes at the price of increased energy consumption compared to both NIN and AlexNet. Hence, for $\alpha=1$ both AlexNet and NIN have a smaller $M$ value. In this case, a machine learner of an embedded Internet-of-Things (IoT) system could use the \textit{Energy-Precision Ratio} to select the most energy efficient architecture. On the contrary, with $\alpha=4$, \emph{i.e.}, when accuracy is being heavily favored over energy efficiency, the $M$ value of VGG-16 is smaller than $M$ value of AlexNet. 

\begin{table}[ht]	
	\centering
	\caption{$M$ metric for different CNN architectures and Energy-per-Image (EPI) values. Network choices could be different for different $\alpha$ values: AlexNet for $\alpha = 1, 2, 3$, VGG-16 for $\alpha = 4$.}
	\label{tab:energy_ratio}
		\small
		\begin{tabular}
			{ccccccc} \toprule
            \multirow{2}{*}{CNN name } & \multirow{2}{*}{Top-5 Error } & \multirow{2}{*}{EPI (mJ) } & \multicolumn{4}{c}{M}  \\ 
            \cline{4-7}
			& & & $\alpha$ = 1 & $\alpha$ = 2& $\alpha$ = 3 & $\alpha$ = 4\\ \hline
			VGG-16 	& 7.30\% & 1178.9	& 86.062	& 6.283	& 0.459	& \textbf{0.033}\\ 
			AlexNet & 17.00\% & 59.3 & \textbf{10.086} & \textbf{1.715}& \textbf{0.291} & 0.050 \\ 
			NIN  & 20.20\% & 89.6 & 18.093 & 3.655 & 0.738 & 0.149 \\\bottomrule
		\end{tabular}	 
\end{table}

With the recent surge of IoT and mobile learning applications, machine learners need to take the energy efficiency of different candidate CNN architectures into consideration, in order to identify the CNN which is more suitable to be deployed on a energy-constrained platform. For instance, consider the case of a machine learner that has to choose among numerous CNNs from a CNN model ``zoo'', with the best error of each CNN readily available. \emph{Which network provides the best trade-off between accuracy for energy spent per image classified?} Traditionally, the main criterion for choosing among CNN architectures has been the data item classification accuracy, given its intuitive nature as a metric. However, there has not been so far an easily interpretable metric to trade-off energy efficiency versus accuracy. 

Towards this direction, we can use our proposed model to quickly predict the total energy consumptions of all these different architectures, and we can then compute the $M$ score to select the one that properly trades off between accuracy for energy spent per image classified. We postulate that the \textit{Energy-Precision Ratio}, alongside our predictive model, could therefore be used as a useful aid for machine learners \emph{towards energy-efficient CNNs}.

\subsection{Models on other platforms}
\label{subsec:models on other platforms}
Our key goal is to provide a modeling framework that could be flexibly used for different hardware and software platforms. To comprehensively demonstrate this property of our work, we extend our evaluation to a different GPU platform, namely Nvidia GTX 1070, and another Deep Learning software tool, namely Caffe by \cite{jia2014caffe}.

\subsubsection{Extending to other hardware platforms: Tensorflow on Nvidia GTX 1070}
We first apply our framework to another GPU platform, and more specifically to the Nvidia GTX 1070 with 6GB memory. We repeat the runtime and power data collection by executing Tensorflow, and we train power and runtime models on this platform. The layer-wise evaluation results are shown in Table \ref{tab:1070_layer}. For these results, we used the same polynomial orders as reported previously for the TensorFlow on Titan X experiments. Moreover, we evaluate the overall network prediction for runtime, power, and energy values and we present the predicted values and the prediction error (denoted as Error) in Table~\ref{tab:whole_model_runtime-tf-1070}. Based on these results, we can see that our methodology achieves consistent performance across different GPU platforms, thus enabling a scalable/portable framework from machine learning practitioners to use across different systems.

\begin{table}[ht]	
	\centering
	\caption{Runtime and power model for all layers using TensorFlow on GTX 1070.}
	\label{tab:1070_layer}
		\small
		\begin{tabular}
			{l|ccc|ccc} \toprule
			\multirow{2}{*}{Layer type } & \multicolumn{3}{c}{Runtime} & \multicolumn{3}{|c}{Power} \\ \cline{2-7}
            & Model size & RMSPE & RMSE (ms) & Model size & RMSPE & RMSE (W)\\ \hline
			Convolutional & 10 & 57.38 \% & 3.5261  & 52 & 10.23\%	&9.4097\\ 
			Fully-connected & 18 & 44.50\% & 0.4929 & 23 & 7.08\%	& 5.5417\\ 
			Pooling & 31 & 11.23\% & 0.0581 & 40 & 7.37\% & 5.1666 \\ \bottomrule
		\end{tabular}	 
\end{table}

\begin{table}[ht]	
	\centering
	\caption{Evaluation of \NP on CNN architectures using TensorFlow on GTX 1070.}
	\label{tab:whole_model_runtime-tf-1070}
		\small
		\begin{tabular}
			{l|cc|cc|cc} \toprule
			\multirow{2}{*}{CNN name} & \multicolumn{2}{c}{Runtime} & \multicolumn{2}{|c|}{Power} & \multicolumn{2}{c}{Energy} \\ \cline{2-7}
            &  Value (ms) & Error &  Value (W) & Error &  Value (J) & Error\\ \hline
			AlexNet &  44.82 & 17.40\%  & 121.21	& -2.92\% & 5.44 & 13.98\% \\
			NIN &  61.08 & 7.24\% &  120.92	& -4.13\% &  7.39	& 2.81\% \\ \bottomrule
		\end{tabular}	 
\end{table}

\subsubsection{Extending to other machine learning software environments: Caffe}

Finally, we demonstrate the effectiveness of our approach across different Deep Learning software packages and we replicate our exploration on Caffe. To collect the power and runtime data for our predictive models, we use Caffe's supported mode of operation for benchmarking, namely \texttt{time}. While this functionality benchmarks the model execution layer-by-layer, the default Caffe version however reports only timing. To this end, we extend Caffe's \texttt{C++} codebase and we incorporate calls to Nvidia's \texttt{NVML C++ API} to report power values. 

We present the per-layer accuracy for runtime and power predictions in Table~\ref{tab:caffe-1070}. Furthermore, we evaluate our model on the AlexNet and NIN networks in Table~\ref{tab:whole_model_runtime-caffe-gtx}. Please note that the execution of the entire network corresponds to a different routine under the Caffe framework, so a direct comparison is not possible. We instead compare against the Equations~\ref{eq:runtime_comparison}-\ref{eq:energy_comparison} as in the previous subsection. Same as for the TensorFlow case before (Table~\ref{tab:1070_layer}), we observe that in this case as well the runtime predictions exhibit a larger error in this platform when executing on the GTX1070 system. This is to be expected, since the GTX 1070 drivers do not allow the user to clock the GPU to a particular frequency state, hence the system dynamically selects its execution state. Indeed, in our collected datapoints we observed that Caffe's (and TensorFlow's previously) higher variance in the runtime values. To properly capture this behavior, we experimented with regressors other that second and third degree polynomials. In these results for Caffe in particular, we select linear models since they provided a better trade off between training error and overfitting.

\begin{table}[ht]	
	\centering
	\caption{Runtime and power model for all layers using Caffe on GTX 1070.}
	\label{tab:caffe-1070}
		\small
		\begin{tabular}
			{l|ccc|ccc} \toprule
			\multirow{2}{*}{Layer type } & \multicolumn{3}{|c|}{Runtime} & \multicolumn{3}{|c}{Power} \\ \cline{2-7}
            & Model size & RMSPE & RMSE (ms) & Model size & RMSPE & RMSE (W)\\ \hline
			Convolutional & 32 & 45.58 \% & 2.2301  & 32 & 6.19\%	& 11.9082\\
			Fully-connected & 18 & 48.41 \% & 0.6626 & 18 & 8.63\%	& 8.0291 \\ 
			Pooling & 30 & 37.38 \% & 0.1711 & 26 & 6.72 \% & 11.9124 \\ \bottomrule
		\end{tabular}	 
\end{table}

\begin{table}[ht]	
	\centering
	\caption{Evaluation of our model on CNN architectures using Caffe on GTX 1070.}
	\label{tab:whole_model_runtime-caffe-gtx}
		\small
		\begin{tabular}
			{l|cc|cc|cc} \toprule
            \multirow{2}{*}{CNN name} & \multicolumn{2}{|c|}{Runtime} & \multicolumn{2}{|c|}{Power} & \multicolumn{2}{|c}{Energy} \\ \cline{2-7}
            &  Value (ms) & Error &  Value (W) & Error &  Value (J) & Error\\ \hline
			AlexNet &  51.18 & -31.97\%  & 107.63	& -5.07 \% & 5.51 & 35.42\% \\
			NIN &  76.32 & 0.36 \% &  109.78 & -8.89\% &  8.38	& 8.56\% \\ \bottomrule
		\end{tabular}	 
\end{table}

\subsection{Discussion}
\added{It is important to note that overhead introduced by \NP is not significant. More specifically, \NP needs to collect datasets to train the models, however, the overhead for training is very small, e.g., around 30 minutes for GPU Titan X. This includes data collection (under 10 minutes) and model training (less than 20 minutes). The process is done once for a new platform. This overhead can be easily offset if the CNN architecture search space is large. Even if machine learners only evaluate a few CNN architectures, \NP can still provide the detailed breakdown with respect to runtime, power, and energy to identify bottlenecks and possible improvement directions.}

\section{Conclusion}
\label{sec:conclusion}
With the increased popularity of CNN models, the runtime, power, and energy consumption have emerged as key design issues when determining the network architecture or configurations. In this work, we propose \NP, the first holistic framework to provide an accurate estimate of power, runtime, and energy consumption. The runtime model of \NP outperforms the current state-of-the-art predictive model in terms of accuracy. Furthermore, \NP can be used to provide an accurate breakdown of a CNN network, helping machine learners identify the bottlenecks of their designed CNN models. Finally, we assess the accuracy of predictions at the network level, by predicting the runtime, power, and energy of state-of-the-art CNN configurations. \NP achieves an average accuracy of $88.24\%$ in runtime, $88.34\%$ in power, and $97.21\%$ in energy. As future work, we aim to extend our current framework to model the runtime, power, and energy of the networks with more sophisticated parallel structures, such as the ResNet network by \cite{he2016deep}.

\textbf{Acknowledgments}: This research was supported in part by NSF CCF Grant No. 1514206 and NSF CNS Grant No. 1564022.


\small
\bibliography{acml17}






\end{document}